\documentclass[final]{cvpr}

\usepackage{times}
\usepackage{epsfig}
\usepackage{graphicx}
\usepackage{amsmath}
\usepackage{amssymb}
\usepackage{amsmath}
\usepackage[utf8]{inputenc} % allow utf-8 input
\usepackage[T1]{fontenc}    % use 8-bit T1 fonts
\usepackage[pagebackref=true,breaklinks=true,letterpaper=true,colorlinks,bookmarks=false]{hyperref}       % hyperlinks
\usepackage{url}            % simple URL typesetting
\usepackage{booktabs}       % professional-quality tables
\usepackage{amsfonts}       % blackboard math symbols
\usepackage{nicefrac}       % compact symbols for 1/2, etc.
\usepackage{microtype}      % microtypography
\usepackage{bm}
\usepackage{sidecap}
\usepackage{graphicx}
\usepackage{subcaption}
\usepackage{float}
\usepackage{lipsum}
\usepackage{multicol}
\usepackage{multirow}
\usepackage{wrapfig}
\usepackage[bottom]{footmisc}
\sidecaptionvpos{table}{c}

\usepackage{colortbl, xcolor}

\usepackage[pagebackref=true,breaklinks=true,colorlinks,bookmarks=false]{hyperref}

\def\cvprPaperID{2328} % *** Enter the CVPR Paper ID here
\def\confYear{CVPR 2021}

\begin{document}

%%%%%%%%% TITLE
\title{Learning Goals from Failure} %Video Representations of Goals Emerge from Watching Failure}

\newcommand{\cv}[1]{{\color{green}[Carl says: #1]}}
\newcommand{\de}[1]{{\color{red}[Dave says: #1]}}

\author{Dave Epstein and Carl Vondrick\\Columbia University\\
\href{https://aha.cs.columbia.edu}{aha.cs.columbia.edu}
}

\maketitle

% \begin{figure*}[t]
% \includegraphics[width=0.6\linewidth]{figs/teaser.pdf}
% \caption{\textbf{Oops!} Did this person intend for this action to happen, or was it an accident? In this paper, we introduce a video dataset of accidental actions and models to localize and predict them.}
% \label{fig:teaser} 
% \end{figure*}
\enlargethispage{-6cm} % -6.6 with anon authors
\noindent\begin{picture}(0,0)
\put(0,-423){\begin{minipage}{\textwidth}
\centering
\includegraphics[width=\linewidth]{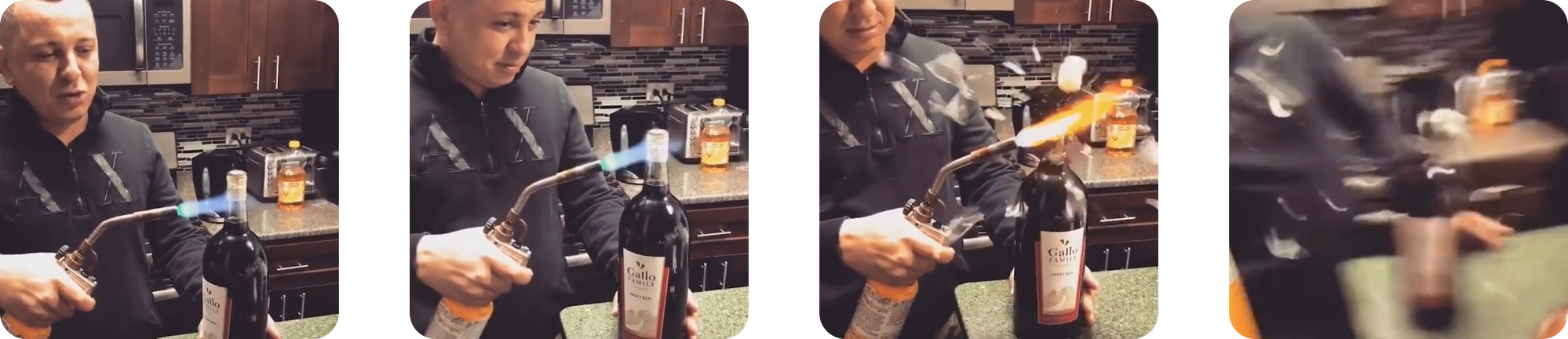}
% \captionof{figure}{\textbf{Intentional versus Unintentional Action:} Did this person intend for this action to happen, or was it an accident? In this paper, we introduce a large in-the-wild video dataset of intentional and unintentional actions. However, due to the realistic diversity of actions, manually annotating human activities is challenging. We propose several self-supervised representation learning approaches for predicting intentionality from video.}
\captionof{figure}{\textbf{What are they doing?} While just the action is observable (heating the bottle), we still predict the goal behind the action (to open the bottle). In this paper, we learn from failure examples to learn representations of goals in video.}
\label{fig:teaser}
\end{minipage}}
\end{picture}%

\begin{abstract}
We introduce a framework that predicts the goals behind observable human action in video. Motivated by evidence in developmental psychology, we leverage video of unintentional action to learn video representations of goals without direct supervision.
Our approach models videos as contextual trajectories that represent both low-level motion and high-level action features. 
Experiments and visualizations show our trained model is able to predict the underlying goals in video of unintentional action. We also propose a method to ``automatically correct'' unintentional action by leveraging gradient signals of our model to adjust latent trajectories. Although the model is trained with minimal supervision, it is competitive with or outperforms baselines trained on large (supervised) datasets of successfully executed goals, showing that observing unintentional action is crucial to learning about goals in video.
\end{abstract}
\vspace{-1em}
\section{Introduction}

%  lay the groundwork necessary for acquiring skills they will rely on for the rest of their lives

Goal-directed action is all around us. Even though Figure \ref{fig:teaser} shows a person performing an unconventional action (heating a wine bottle with a blowtorch),  we cannot help but to perceive the action as rational in the context of the goal (to open the bottle). 
%Recently, computer vision has made tremendous progress in action recognition  \cite{carreira2017quo, ji2019action}, which classifies how a person acted, but not why they acted. %However, the resulting video representations do not discriminate the underlying goals of action. 

Predicting the goal of action may seem challenging because future goals are not directly observable in video. However, in a series of papers, development psychologists  Amanda Woodward and Michael Tomasello demonstrated that children reason about goals before their second birthday \cite{tomasello2009usage, woodward2009infants}, and this reasoning plays a key role in rapid development of communicative skills \cite{tomasello2005understanding} and mental representations of the world \cite{barresi1996intentional}. Despite the relative ease of this task for children,  machine recognition of goals has remained challenging.

% \de{FIGURE: teaser + frame problem} Newborn babies find themselves immediately thrust into a social world overwhelmed by new visual stimuli, with no world knowledge whatsoever. Yet, in their first year, infants manage to lay the groundwork necessary for acquiring skills they will rely on for the rest of their lives \cite{tomasello2009usage}. An understanding of intention is central to this groundwork, since it allows children to learn to communicate rapidly \cite{tomasello2005understanding} and to develop rich mental representations of the world \cite{barresi1996intentional} (what some call ``theory of mind''). 

The hypothesis underlying this paper is that examples of \emph{failure} are key missing pieces in action recognition systems. 
Without observing unintentional action, we cannot expect models to discriminate goals from actions. Examples demonstrating unintentional action are necessary to decouple these two notions, separating between the visible action and the latent goals. As Efros has been telling us all along, it is all about the data \cite{hays2007scene}, and negative data doubly so \cite{wu2017sampling}.%Similar to how a child learns about goals by experiencing failure,

% \begin{figure}[t]
% \includegraphics[width=\columnwidth]{teaser.png}
% % \vspace{-1em}
% \caption{\textbf{What is this person's goal?} Although only the action is observable, we are still able to predict the goal behind the action (to open the bottle). In this paper, we introduce a model to learn video representations that encode goals as latent action trajectories.}
% \label{fi

\enlargethispage{-6cm} % -6.6 with anon authors

The main observation behind our approach is that natural video will contain abundant and rich examples of both intentional and unintentional action \cite{oops}, which we can leverage for learning. In our model, video is represented as a trajectory, and goals are encoded as the path for the trajectory. Given examples of videos with variable success, we present a model that learns goal-oriented video representations by discriminating between success and failure. Our model captures both motion and relational features through an attention-based transformer architecture, allowing end-to-end training.

%We present a model that leverages examples of both intentional and unintentional action  to learn goal-oriented representations of video.
%Our approach learns a trajectory representation of action, and encodes goals as the path of the trajectory. We input entire videos to our model by first dividing them into short clips,
%which are run through a 3D CNN to learn low-level motion features. We then pass the motion features
%into a Transformer model, which models relations between different periods in its input, and thus represents the entire action as a path through latent space. The whole model is trained from scratch in an end-to-end manner.

\begin{figure*}[t]
\centering
\begin{minipage}[c]{0.6\textwidth}
\includegraphics[width=\textwidth]{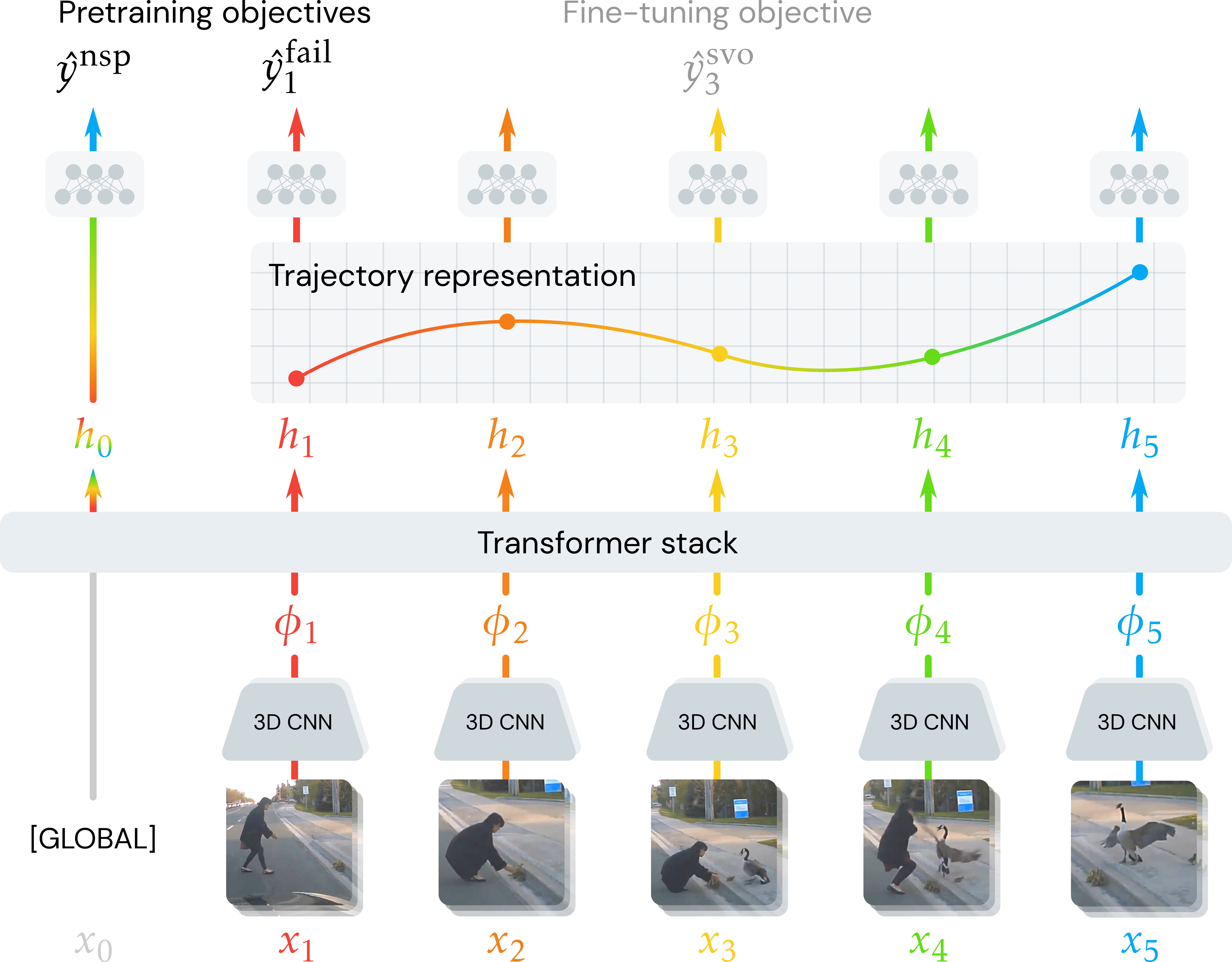}
% \vspace{-0.5em}
 \end{minipage}\hfill
\begin{minipage}[c]{0.39\textwidth}
\caption{\textbf{Learning goal-oriented video representations:} We show an overall view of our approach. First, we embed short clips using a 3D CNN to represent short-term motion features. Then, we run the sequence of CNN embeddings through a stack of Transformers, where they interact with each other to finally form a context-adjusted latent action trajectory. The model is trained end-to-end from scratch, with intentionality and temporal coherence losses (depicted top-left). Points along the resultant trajectory are decoded with linear projections into various spaces (top-middle).}
\label{fig:arch} 
 \end{minipage}
\end{figure*}

Our experiments show that failure data is crucial for learning representations of goals. 
We evaluate our model on three goal prediction tasks. First, we experiment on detecting unintentional action in video, and we demonstrate strong performance over baselines on this task. Second, we evaluate the representation at predicting goals with minimal supervision, which we characterize as structured categories consisting of subject, action, and object triplets. Lastly, we use our representation to automatically ``correct'' unintentional action  and decode these corrections by retrieving from other videos or generating categorical descriptions. 

Our main contribution is an approach that, training on data of unintentional action, learns a goal-directed representation of videos. We show that our model often captures the latent goals behind observed action, performing on par with or better than supervised models trained on large labeled datasets of only intentional action. We also introduce a method to find minimal adjustments to the path and ``automatically correct'' unintentional action in video. The remainder of this paper will describe this approach in detail. Code, data, and models will be available.

\section{Related Work}

\textbf{Recognizing action in video:} Previous work explores many different approaches to recognizing action in video. Earlier directions develop hand-designed features to process spatio-temporal information for action recognition \cite{laptev2005space,klaser2008spatio,wang2011action,pirsiavash2014parsing}. Popular deep learning architectures for images were extended to operate directly on video by modeling time as a third dimension \cite{hara2018can,carreira2017quo,simonyan2014two, luvizon20182d,ji2019action}. To deal with variable-length or long video input, previous work frequently takes one of two approaches: pooling or recurrent networks. However, pooling loses spatial and/or temporal connections between different moments of video. Since recurrent networks are sequential, they require selecting important video features ahead of time, without viewing full context. RNNs are also known to struggle to connect between far-apart inputs, which creates significant challenges in modeling long-term video. \cite{sun2019contrastive} is most similar to our approach, since they also run clips through 3D CNNs and Transformers, but they freeze 3D CNNs and train on a ``masked video modeling" task, ultimately discarding contextually learned temporal dynamics across videos since their goal is to learn information useful for an effective cross-modal representation. To address these drawbacks, we propose a 3D-CNN-Transformer model which allows for short-term, granular motion detection combined with a long-term action representation, trained end-to-end from scratch. 
% We introduce a temporal coherence loss inspired by this previous work. \cite{pirsiavash2014assessing,doughty2018s,Parmar_2017_CVPR_Workshops} explore action quality evaluation, most often requiring significant manual labels. We view quality through the lens of intentional and automatically correct unintentional action without requiring labels.

% Several methods also explore using the incidental statistics of videos to learn self-supervised representations \cite{han2019video,misra2016shuffle,fernando2017self,wei2018learning,jayaraman2016slow}. We introduce a temporal coherence loss inspired by this previous work. 

\textbf{Learning about intention:} Evidence in developmental psychology quantifies why humans perceive intention \cite{barresi1996intentional}, how we perceive it \cite{woodward2001infants, woodward2009emergence, woodward2009infants}, when we begin to do so \cite{meltzoff1995understanding, meltzoff1999toddlers}, and what allows us to infer the goals behind others' behavior \cite{shultz1980development}.
Early work in computer vision has investigated assessing the quality of action execution
\cite{pirsiavash2014assessing,doughty2018s,Parmar_2017_CVPR_Workshops}, which our work builds upon. However, we view quality from a goal-directed perspective and automatically correct unintentional action with minimal supervision.
%While these questions have been studied in early stages of child development, the same abilities have remained a challenge for machines in unconstrained situations. One possible reason for this is a lack of realistic data, or data depicting unintentional action.
We take advantage of signals in unconstrained video collections of both intentional and unintentional action \cite{oops} to learn about goals from video.

\textbf{Leveraging adversarial attacks:} We use adversarial gradients  \cite{goodfellow2014explaining, kurakin2016adversarial} to find adjustments to learned video representations which ``auto-correct'' unintentional action back onto the manifold of intentional action. Previous work studied adversarial attacks in steganography \cite{hayes2017generating, zhu2018hidden}, software bug-finding \cite{she2019neuzz}, generating CAPTCHAs \cite{von2003captcha} to fool modern deep nets \cite{osadchy2017no}, generating interesting images \cite{simonyan2013deep}, creating real-world 3D objects that trick neural networks \cite{zhou2018invisible, athalye2017synthesizing}, and in training models more robust to test-time adversarial attacks \cite{miyato2015distributional, goodfellow2014explaining, miyato2016adversarial}. \cite{jahanian2019steerability}  extend this concept to generative models, setting a new image output as a target label and perturbing latent space. In video, \cite{jiang2019black, wei2018transferable} introduce various methods to fool action recognition networks, often on a 3D CNN backbone. We instead utilize adversarial attacks to manipulate and correct unintentional action.

% \cv{I think you can delete this paragraph.}
% \textbf{Automatic error correction:} Machine learning systems that can automatically fix mistakes in input have been highly studied in natural language processing. \cite{rozovskaya2016grammatical,gamon2008using,schmaltz2017adapting} propose deep neural net methods for correcting spelling and grammar issues in text input with varying levels of supervision, building on classical rule-based approaches ({\em e.g.} \cite{kukich1992techniques, van1988triphone}). In the image domain, texture-based \cite{barnes2009patchmatch} or deep learning \cite{yeh2016semantic,pathak2016context} methods can be used to remove undesired content from image regions and sensibly fill in the blank space. Extending these semantic notions of ``fixing mistakes'', we train a model to automatically correct unintentional action in video with minimal supervision.

\section{Unintentional Action and Goals Dataset}

\begin{figure}[t]
\centering
\includegraphics[width=0.8\columnwidth]{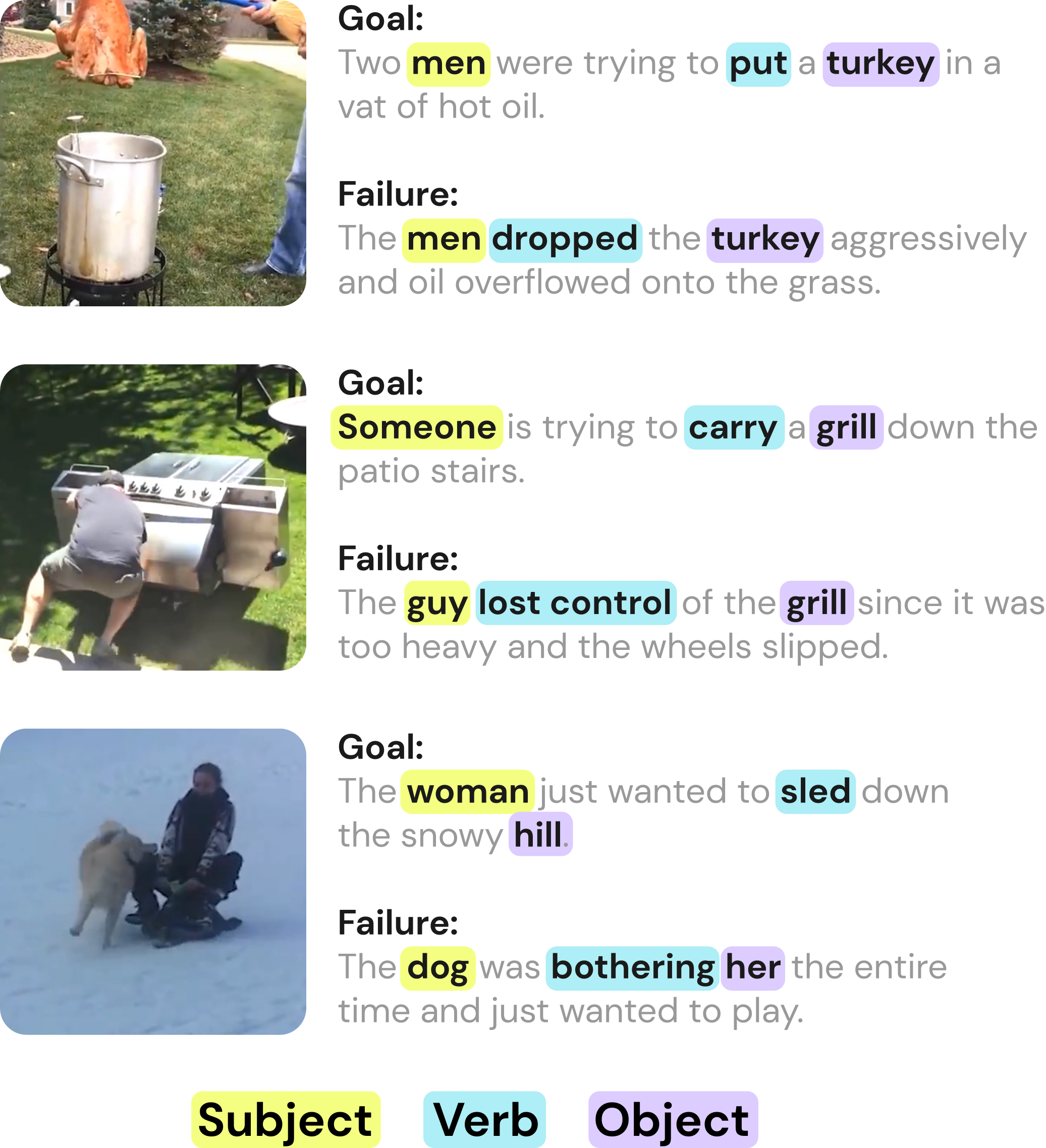}
% \vspace{-0.5em}
\caption{\textbf{Labeling goals and failures in video:} To evaluate our representation, we annotate the Oops! dataset with short sentences describing the goals and failures. We extract subject-verb-object triples and train a decoder on learned representations. The intentional and unintentional action in the dataset span a diverse range of categories.}
\label{fig:dataset} 
\end{figure}

Similar to how children learn about goals by perceiving failed attempts at executing them \cite{meltzoff1999toddlers}, we hypothesize that examples of failure are crucial for learning to discriminate between action and goal. Without observing unintentional action, models can not learn the pattern discriminating action and intention. We build on the Oops! dataset \cite{oops}, which is a large collection of videos containing intentional and unintentional action, to train and evaluate our models. Videos in this dataset are annotated with the moment at which action becomes unintentional.
% \footnote{In addition to the ground truth annotations provided by \cite{oops}, we run their pretrained model on the unlabeled portion of the training set and collect pseudo-ground-truth, which we found improves performance. This can be seen as a weak instantiation of born-again networks \cite{furlanello2018born}.} 
Figure \ref{fig:dataset} shows some example frames.  We also use the Kinetics dataset \cite{kinetics} to evaluate models, since it contains a wide range of successful actions. 

We would like to learn a representation of goals that only requires visual information to train. However, evaluating trained models and probing them for an understanding of goals requires gathering labels of goals. Therefore, we expand \cite{oops} with textual descriptions of goals and failures in the dataset, and use these annotations to evaluate our (trained, frozen) model in comparison to other representations.

% \subsection{Labeling goals in Oops!}
\subsection{Goal and Failure Annotation}
\label{sec:lbl_svos}
% \textbf{Goal Annotation:}
% We train with indirect supervision, but we require goal labels for evaluation.
Established action datasets in computer vision \cite{gu2018ava, li2020ava} contain annotations about person and object relationships in scenes, but they do not directly annotate the goal, which we need for evaluation of goal prediction. We collect unconstrained natural language descriptions of a subset of videos in the Oops! dataset (4675 training videos and 3404 test videos), prompting Amazon Mechanical Turk workers\footnote{with $>10k$ approvals at a $\geq 99\%$ rate} to answer ``What was the goal in this video?'' as well as ``What went wrong?''. We then process these sentences\footnote{Using the Spacy.io natural language library} to detect lemmatized subject-verb-object triples, manually correcting for common constructions such as ``tries to X'' (where the verb lemma is detected as ``try'', but we would like ``X'').  The final vocabulary contains 3615 tokens. Figure \ref{fig:dataset} shows some example annotations. Detailed statistics for processed SVO triples are provided in the Supplementary Material. We use SVO triples to evaluate the video representations.

\section{Method}

In this section, we introduce our framework to learn goal-oriented trajectory representations of video. Our method accepts as input sequences of video input depicting intentional and/or unintentional action, and learns to represent these sequences as latent trajectories, from which intentionality of action is predicted. We show in Section \ref{sec:exp} that, having observed unsuccessful action as well as successful, our trained model learns trajectories which capture the goals latent in the input video.

\subsection{Visual Dynamics as Trajectories}

A common approach to representing video data is to run each clip through a convolutional network and combine clip representations by pooling to run models on entire sequences  \cite{feichtenhofer2019slowfast, han2019video, gao2019listen,xu2017r}. However, these methods do not allow for connections between different moments in video and cannot richly capture temporal relationships, which give rise to goal-directed action. While recurrent networks \cite{hochreiter1997long} are more expressive, they require compressing history into a fixed-length vector, which forces models to select relevant visual features without viewing full context and makes reasoning about connections between different moments difficult, especially when they are far apart. 
% Further, since RNNs must process data sequentially, they are forced to select relevant visual features without viewing full context.
% For example, consider a video with two clips: in the first, X, and in the second Y. Pooling the embeddings for these clips together could not possibly yield the correct representation, which is likely that Z. \de{TODO find example}

Temporal streams of visual input are highly contextual with both short- and long-term dependencies. We will represent video as a contextually-adjusted trajectory of latent representations in a learned space. Figure \ref{fig:arch} illustrates this architecture, which has both a motion and action level:

\textbf{Motion Level:} First, we separate video into short clips (or tokens) in order to make initial motion-level observations. 
Let $x$ be a video, and $x_i$ be a video clip centered at time $i$. We estimate the motion-level features $\phi_i = f(x_i)$ where $f$ is a 3D CNN \cite{3dcnn}.

\textbf{Action Level:} Second, we model relations between $\phi_i$ to construct a contextual trajectory $h_i = g(\phi_i)$  where $g$ is the Transformer \cite{transformer}.   The Transformer accepts as input a sequence of motion-level representations $\{\phi_i\}_{i=1}^n$, repeatedly performs self-attention among them, in the same spirit as the forward pass of a graph neural network, with video clips as nodes \cite{wu2020comprehensive}. The output of the Transformer is a final latent path $\{h_i\}_{i=1}^n$. Since the self-attention operation can incorporate contributions from both nearby and far away moments in its representations for each clip, the Transformer is well-suited to modeling higher-level connections between the atomic actions recognized at the motion level. The Transformer's output $\{h_i\}_{i=1}^n$ can then be applied in different downstream tasks.
%(Section \ref{sec:acorr} and Section \ref{sec:decode}).

%While previous work \cite{EXPERT, lu2019vilbert, li2019visualbert, tan2019lxmert, su2019vl} has considered 2D CNN Transformers, these models often struggle to capture low-level, granular details in action \cite{EXPERT}. In this paper, 

%To our knowledge, we are the first to leverage 3D CNNs in combination with Transformers to represent video.

\subsection{Learning with Indirect Supervision}
\label{sec:learning}
% An immediate obstacle in attempting to train models to learn such trajectories is that the space of goals is vast, making it challenging to label exhaustively. Goals can vary widely, there is no ground truth, and they exhibit hierarchical structure.
We learn the representation with weak, indirect supervision that is accessible at large scales. This supervision is also truer to how humans learn about intention, since we do not require labeled action semantics, but do often receive environmental cues about whether others' action is intentional or not \cite{colle2007childrens}. %We use the moment when action becomes unintentional in a video, which provides examples of failure. We also use a self-supervised loss to induce robust representations of temporal relationships.
We use the following two objectives for learning:

\textbf{Action Intentionality:} We train the model to temporally localize when action is unintentional. We assume that the video frame where the action shifts from intentional to unintentional is labeled \cite{oops}, and note that these labels are a significantly weaker form of supervision than semantic action categories. For each video clip $x_i$, we set the target $y^\text{fail}_i \in \{0,1,2\}$ according to whether the labeled frame happens before, during, or after the clip $x_i$. The model estimates $\hat{y}^\text{fail}_i = \textrm{softmax}(w_1^T h_i)$ with a linear projection where $w_1$ is a jointly learned projection matrix to $\mathbb{R}^3$. We train with a cross-entropy loss between $\hat{y}^\text{fail}$ and $y^\text{fail}$ where the class weight is set to the inverse frequency of the class label to balance training. We label this loss $\mathcal{L}^\text{fail}$.

%a linear layer $f_{intention}$ with activation function (we use GELU \cite{hendrycks2016gaussian}) to yield a three-dimensional logits vector $p_i$ that is followed by a softmax and trained with the cross-entropy loss $\mathcal{L}_{intention}(p,y) = - w_y \log \frac{\exp(p_y)}{\sum_{y^\prime \in \{0, 1, 2\}} \exp(p_{y^\prime})}$. We set the class weight $w_y$ to the inverse frequency of the class label in training to balance training.  

\textbf{Temporal Consistency:}\label{sec:coherence} We also train the model to learn temporal dynamics with a self-supervised consistency loss \cite{han2019video,misra2016shuffle,fernando2017self,wei2018learning,jayaraman2016slow,BERT}.
Let $y^\text{nsp} = 1$ indicate that the sequence is consistent. We predict whether the input sequence is temporally consistent with $\hat{y}^\text{nsp} = \sigma(w_2^T h_0)$
where $w_2$ is a jointly learned projection to $\mathbb{R}$. We train with the binary cross-entropy loss between $y^\text{nsp}$ and $\hat{y}^\text{nsp}$. We label this loss $\mathcal{L}^\text{nsp}$ (next sequence prediction). This loss encourages the model to learn longer-term patterns in human action.

We create inconsistent sequences as follows:
For each video sequence in the batch, we bisect the sequence into two parts at a random index with probability $p_\text{split} = 0.5$. 
% ($p_{split} < 1$, since at test-time we would like to input one entire, unsegmented video)
For these sequences, we perturb the video segments with probability $p_\text{perturb} = 0.5$. When perturbing,  we swap the order of the two sequences with probability $p_\text{swap} = 0.3$, otherwise we pick a randomly sized subsequence from another video sequence in the batch to replace one of the two segments. 

A large line of recent and concurrent work has tackled the problem of self-supervised representation learning in video (\eg \cite{han2019video, han2020memory, han2020self, misra2016shuffle, fernando2017self}). Our paper focuses on the value of training on data of unintentional action to learn goals, and we use the self-supervised temporal consistency loss to encourage our model to reason about longer sequences of action, especially useful for the automatic correction of unintentional action demonstrated in Section \ref{sec:ac_eval}. Other self-supervised losses could be incorporated into our framework to serve the same purpose.

\textbf{Training:} To train our model, we set the overall loss as $\mathcal{L} = \mathcal{L}^\text{fail} + \lambda \mathcal{L}^\text{nsp}$, where $\lambda$ is a hyperparameter controlling the importance of the coherence loss. We set $\lambda = 0.5$ to balance the magnitudes of the losses.
% We sample sequences of one-second long clips, run each clip $x_i$ through the motion-level 3D CNNs then pass all outputs through the Transformer stack, and calculate the gradients. We optimize the loss with stochastic gradient descent. 
%We randomly bisect and perturb video sequences to train on temporal coherence.
% At inference time, we run entire continuously-sampled videos through our model.
%For additional details, please see the Supplementary Material.

%We train a linear projection on $h_0$ (the embedding of a special token used to represent the whole video) to make the binary prediction on whether the two sequences are temporally consistent , using softmax and cross-entropy loss. 

\section{Experiments}\label{sec:exp}

\subsection{Experimental Setup}
\label{sec:decode}

% We use the standard train/test split from \cite{oops} to pre-train, fine-tune, and evaluate models. After learning the trajectory representation, we need to decode it into a space that we can interpret and evaluate. To estimate decoders, we freeze the entire base model up to the $h$ vectors. We consider two decoder modalities. 

% Our model and all baselines are trained on the action intentionality loss (Section \ref{sec:learning}) using the train/test split in \cite{oops}. To evaluate, we freeze entire models up to their output representations (denoted $h$ for our model), and implement various decoders that take model representations as input.

\textbf{Baselines:} We evaluate the 3D CNN from \cite{oops} which is trained from scratch on the action intentionality loss (Section \ref{sec:learning}). We also evaluate a 3D CNN pre-trained on Kinetics action recognition, which is frozen unless indicated otherwise. The 3D CNN trained on Kinetics is the current state of the art in video representation learning when transferred to many downstream tasks, and represents a high-water mark for performance when training only on intentional action. Further, to fairly compare the Transformer layer to 3D CNNs which take in one short clip only, we pool 3D CNN predictions locally with neighboring predictions such that both methods have the same effective temporal receptive field.

We evaluate our learned representations by freezing them and then decoding them via retrieval as well as goal and failure prediction.

\textbf{Retrieval:} We perform nearest-neighbor retrieval among one-second long clips in the test sets for the Oops! and Kinetics datasets. While we do not learn a representation using Kinetics data, we include a subset of Kinetics (of the same size as the Oops! validation set) in retrieval, to see if auto-corrected actions match with successfully executed goals in Kinetics rather than failed attempts (see Section \ref{sec:ac_eval}). This decoder maintains a lookup table of all clip representations and computes the $k$-nearest neighbors from different videos using cosine distance.

\textbf{Categorization:} We also implement a decoder using the textual labels we gathered on the videos. Here, the task is to describe the goals of the input video using the SVO triplets. We train a decoder to predict the {\em main goal} for clips with intentional action (before the onset of failure), and predict {\em what went wrong} for clips with unintentional action, using labels gathered as described in Section \ref{sec:lbl_svos}. The estimated decoder will describe intentional action in video with descriptions of the goal, for example ``athlete wins game'' and not ``throwing ball", which is an action. Unintentional action, in turn, will be described as ``man spills groceries'' instead of a generic action category such as ``walking''. We train a linear layer to output a vector for subject, verb, and object. As ground truth, we use BERT word embeddings \cite{BERT}, calculating scores using dot product and running them through softmax and a cross-entropy loss.

% %\begin{wraptable}{R}{.6\textwidth}
% \begin{table}%[2][t]
% % \begin{table}[tb]
% % \centering
% %\resizebox{0.65\linewidth}{!}{
% \resizebox{\columnwidth}{!}{
% \begin{tabular}{c l|c c | c}
% \toprule
% && \multicolumn{2}{c|}{\textbf{Localization}} & \textbf{Classification} \\
% &\textbf{Method} & \textbf{0.25 sec} & \textbf{1 sec} & \textbf{Accuracy}\\ \midrule
% &Kinetics supervision \cite{carreira2017quo} & 69.2 & 37.8 & 53.6 \\
% &\cite{carreira2017quo} + finetune & \textbf{75.9} & \textbf{46.7} & 64.0 \\  \midrule
% &3D CNN only \cite{oops} & 68.7 & 39.8 & 59.4 \\  \midrule
% \multirow{4}{*}{\rotatebox[origin=c]{90}{Our model}} & Classification only & 64.9 & 33.6 & 73.0 \\
% &   + Pseudo-GT & \textbf{72.4} & \textbf{39.9} & \textbf{77.7} \\
% &   + Coherence loss & 63.2 & 32.4 & 72.1 \\
% &    \quad + Pseudo-GT & 71.8 & 39.6 & \textbf{77.8} \\ \midrule
% & Chance & 25.9 & 6.8 & 33.3 \\ \bottomrule
% %\vspace{-3em}
% \end{tabular}
% }
% % }
% % \vspace{1em}
% \caption{\textbf{Detecting unintentional action:} We evaluate models on classifying and localizing unintentional action. Our model is competitive with Kinetics pretraining despite training from scratch, and outperforms it on classification. \de{may be worth only showing the strongest variant of our model instead of confusing reviewers with unclear ablations}}
% \label{tbl:oops_eval}  
% % \vspace{-2em}
% % \end{table}
% %\end{wraptable}
% \end{table}
%\begin{wraptable}{R}{.6\textwidth}
\begin{table}%[2][t]
% \begin{table}[tb]
% \centering
%\resizebox{0.65\linewidth}{!}{
\resizebox{\columnwidth}{!}{
\begin{tabular}{ l|c c | c}
\toprule
& \multicolumn{2}{c|}{\textbf{Localization}} & \textbf{Classification} \\
\textbf{Method} & \textbf{01 sec} & \textbf{0.25 sec} & \textbf{Accuracy}\\ \midrule
Kinetics \cite{carreira2017quo} finetune & \textbf{75.9} & \textbf{46.7} & 64.0 \\
Kinetics frozen + linear & 69.2 & 37.8 & 53.6 \\
3D CNN only \cite{oops} & 68.7 & 39.8 & 59.4 \\
% Classification only & 64.9 & 33.6 & 73.0 \\
   Our model & \textbf{72.4} & \textbf{39.9} & \textbf{77.9} \\ \midrule
%   + Coherence loss & 63.2 & 32.4 & 72.1 \\
    % \quad + Pseudo-GT & 71.8 & 39.6 & \textbf{77.8} \\ \midrule
 Chance & 25.9 & 6.8 & 33.3 \\ \bottomrule
%\vspace{-3em}
\end{tabular}
}
% }
% \vspace{1em}
\caption{\textbf{Detecting unintentional action:} We evaluate models on classifying and localizing unintentional action on the Oops! Our model is competitive with Kinetics supervised features on unintentional action localization despite training from scratch, outperforming it on three-way classification. Since our model learns how to relate between different moments in time, instead of naively pooling, it is able to make better use of temporal context to solve these tasks.}
\label{tbl:oops_eval}  
% \vspace{-2em}
% \end{table}
%\end{wraptable}
\end{table}

\subsection{Unintentional Action Detection}
\label{sec:oops_eval}

\begin{figure*}[t]
\centering
\includegraphics[width=0.8\textwidth]{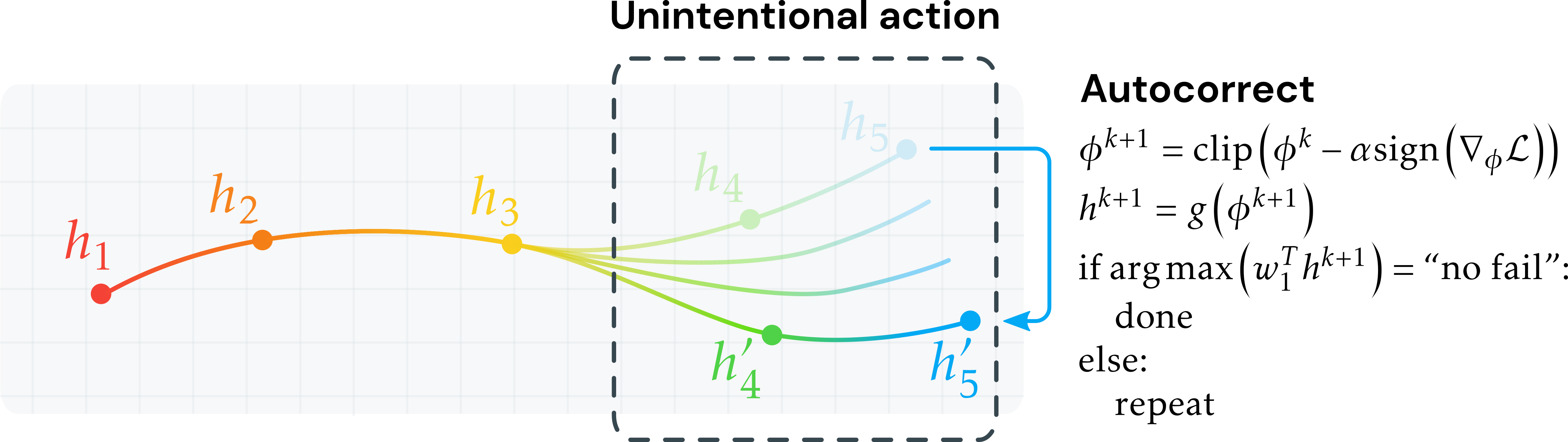}
% \vspace{-0.5em}
\caption{\textbf{Automatically correcting unintentional action:} Starting from an initial trajectory, we use model gradients as a signal to correct the course of points representing unintentional action (Section \ref{sec:acorr}). We evaluate corrected trajectories by decoding them into SVO triples and retrieving nearest neighbors from a databank.}
\label{fig:autocorr}  
\end{figure*}

We evaluate the model at detection and temporal localization when action deviates from its goal. We use labels from  the test set in \cite{oops} as the ground truth. We process videos with our model, sampling continuous one-second clips as tokens, and take the predicted localization as the center of the clip with maximum probability of failure. We also classify each clip according to its label (intentional, transitional, or unintentional). We show results in Table \ref{tbl:oops_eval}. On the former task, our model is competitive with fine-tuning a fully-supervised Kinetics CNN, despite using less data and less supervision. On classification, our network outperforms the Kinetics network by 14\%, showing that representing videos as contextual trajectories is effective.
% \de{TABLE: oops classification and localization accuracy - with baselines from last paper, ablations}

\subsection{Goal Prediction}
\label{sec:svos}

We next evaluate the model at predicting goal descriptions. We train a decoder on the trajectory to read out subject, verb, object triplets. In this task, ground truth is the labeled goal if action is intentional, and the labeled failure if action is unintentional. In training, if sentences have more than one extracted SVO, we randomly select one as ground truth. In testing, we average-pool predictions among all clips with intentional action and unintentional action separately and take the maximum over all sentence SVOs. Each video clip has two pooled predictions: one for video showing intentional action (where ground truth is the labeled goal of the video), and one for video showing unintentional action (where ground truth is the labeled failure). Table \ref{tbl:svo_eval} shows our model obtains better top-1 accuracy on all metrics than baselines, including the Kinetics-pretrained model, and is competitive on top-5 accuracy, highlighting the importance of observing failure for understanding goals.

% \de{TABLE: SVO accuracy table (t1, t5, all3 t1, t5) - with ablations}

% \de{FIGURE: SVO example predictions - pre oops, post oops, post AC}
% \de{FIGURE: correlated neurons?}

\subsection{Completing Goals by Auto-Correcting Trajectories}

%\begin{wraptable}{R}{.6\textwidth}
\begin{table}[t]
 \setlength{\tabcolsep}{3pt}
% \begin{table}[tb]
\vspace{1em}
\centering
% \resizebox{0.55\linewidth}{!}{
\resizebox{\columnwidth}{!}{
\small
\begin{tabular}{l|cc|cc|cc|cc|cc}
\toprule
% && \multicolumn{2}{c|}{\textbf{Localization within}} & \textbf{Classification} \\
& \multicolumn{2}{c|}{\textbf{Subject}} & \multicolumn{2}{c|}{\textbf{Verb}} & \multicolumn{2}{c|}{\textbf{Object}} & \multicolumn{2}{c|}{\textbf{Average}} & \multicolumn{2}{c}{\textbf{All three}} \\
\textbf{Features} & \textbf{R1} & \textbf{R5} & \textbf{R1} & \textbf{R5} & \textbf{R1} & \textbf{R5} & \textbf{R1} & \textbf{R5} & \textbf{R1} & \textbf{R5}\\ \midrule
Kinetics \cite{carreira2017quo} & 26.8 & 72.3 & 27.3 & 52.7 & 36.0 & \textbf{64.6} & 30.0 & \textbf{63.2} & 2.1 & \textbf{16.5} \\
3D CNN \cite{oops}  & 29.4 & 72.7 & 26.4 & 50.4 & 44.7 & 57.9 & 33.5 & 60.3 & 2.9 & 13.9\\
Random & 23.7 & 55.7 & 22.7 & 45.4 & 44.8 & 52.7 & 30.4 & 51.3  & 1.4 & 8.7 \\ \midrule
 Our Model & \textbf{34.3} & \textbf{74.5} & \textbf{29.7} & \textbf{54.2} & \textbf{45.0} & 58.2 & \textbf{36.3} & 62.3 & \textbf{3.3} & 14.4 \\ \midrule
 Chance  & \multicolumn{8}{c|}{0.1} & \multicolumn{2}{c}{<0.1}\\ \bottomrule
\end{tabular}
%}
}
% \vspace{0.6em}
\caption{\textbf{Comparison of Representations:} To evaluate how well representations encode goals, we freeze them and estimate a linear projection to predict labelled subject-verb-object triples in the Oops! validation set. We evaluate top-1 and top-5 recall (R1, R5). By observing sequences of both intentional and unintentional action, our model performs competitively with others trained on large labeled datasets of successful action.
%Since the data is quite sparse (only 18 thousand out of a possible 47 billion SVO combinations appear), predicting correct triples is challenging.
%Although our model is trained with less supervision, our model is competitive with a highly-supervised Kinetics model on the task.
}
\label{tbl:svo_eval}  
% \vspace{-2em}
% \end{table}
%\end{wraptable}
% \vspace{-1em}
\end{table}

\label{sec:acorr}

We would like to use our learned representation in order to infer the goals of people in scenes of unintentional action.
%as infants demonstrate the ability to do \cite{meltzoff1995understanding,skulmowski2015investigating}.
However, since the model is trained with indirect supervision, the trajectories $h$ are not supervised with goal states. We propose to formulate goal completion as a latent trajectory prediction problem. Given an observed trajectory of unintentional action $h$, we seek to find a new, minimally modified trajectory $h^\prime$ that is classified as intentional. By analogy to how word processors auto-correct a sentence, we call this process \textbf{action auto-correct}. We illustrate this process in Figure \ref{fig:autocorr}.

\begin{figure*}[t]
\centering
\includegraphics[width=0.95\textwidth]{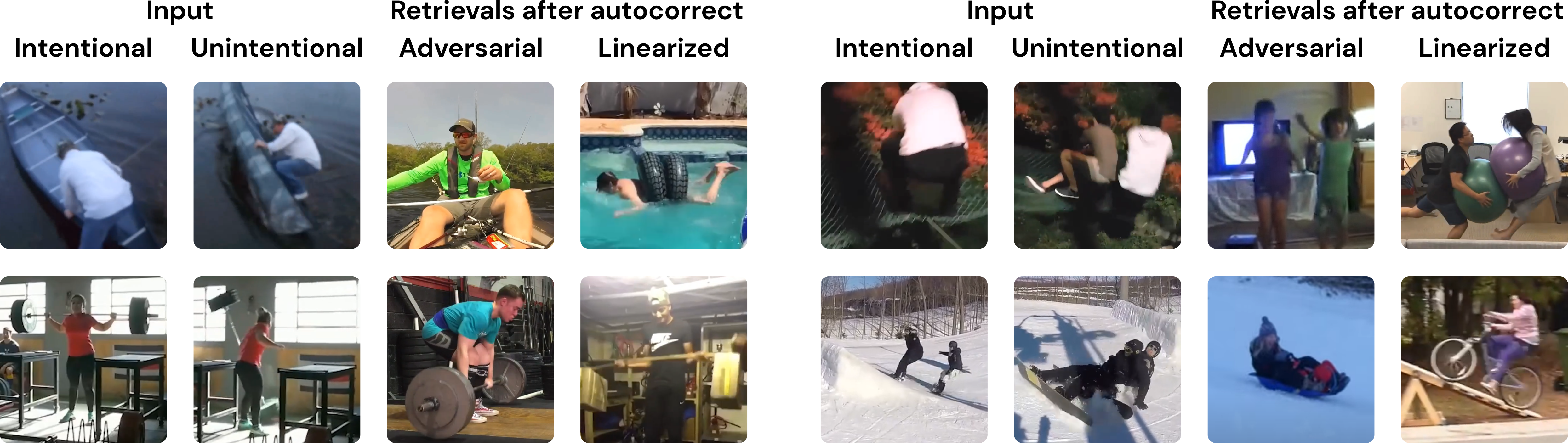}
% \vspace{-0.5em}
\caption{\textbf{Retrievals from Auto-corrected Trajectories:} We show the nearest neighbors from auto-corrected action trajectories, using our proposed method and a linearization baseline. The retrievals are computed across both the Oops! and Kinetics datasets, since Kinetics contains many examples of goals being successfully executed, whereas Oops! focuses on unintentional action.
The corrected representations yield corrected trajectories that are often embedded close to the goal.}
%Our approach retrieves intentional action from the Kinetics dataset with the goals of kayaking, performing a snatch lift, jumping together, and landing a ski trick, respectively (top-to-bottom, left-to-right).
\label{fig:autocorr_retrs} 
\end{figure*}

% \cv{One point of confusion: What is $y_intention$? Is this the same as $y$ in the above paragraphs?}
We find this correction in feature space, not pixel space, to yield interpretable results. We find a gradient to the features $\phi$ that switches the prediction $\hat{y}^\text{fail}_i$ to be the ``intentional'' category for all clips $i$. 
%By perturbing $\phi$ to maximize this classification score, we correct the error and estimate a new trajectory $h^\prime$ that encodes the successful completion of the goal. 

We formulate an optimization problem with two soft constraints. Firstly, we want to increase the classification score of intentional action $\mathcal{L}^\text{fail}$. Secondly, we want the resulting trajectory to be temporally consistent $\mathcal{L}^\text{nsp}$. Without this term, the corrected trajectory is not required to be coherent with the initial part of the original trajectory. We minimize this modified cost function with respect to ${\phi^\prime_{t:T}}$:
\begin{align*}
J = \max\left(0, \mathcal{L}^\text{nsp}_{y=1}(\phi^\prime) - \mathcal{L}^{\text{nsp}}_{y=1}(\phi)\right)
+ \lambda\sum_i \mathcal{L}^\text{fail}_{y=0}(\phi_i^\prime)
\end{align*}
where $\mathcal{L}$s are the original loss functions but with target labels $y^\text{fail}$ overridden to be the intentional class, and $\lambda = 2$ is a scalar to balance the two terms.
We only modify $\phi$ on the clips which the model classifies as unintentional in the first place, which we denote $\phi^\prime_{t:T}$. The coherence loss is also truncated by its original value, causing the optimization to favor a trajectory that is no less temporally coherent than the original one.

\begin{figure*}[t]
\centering
\includegraphics[width=0.95\textwidth]{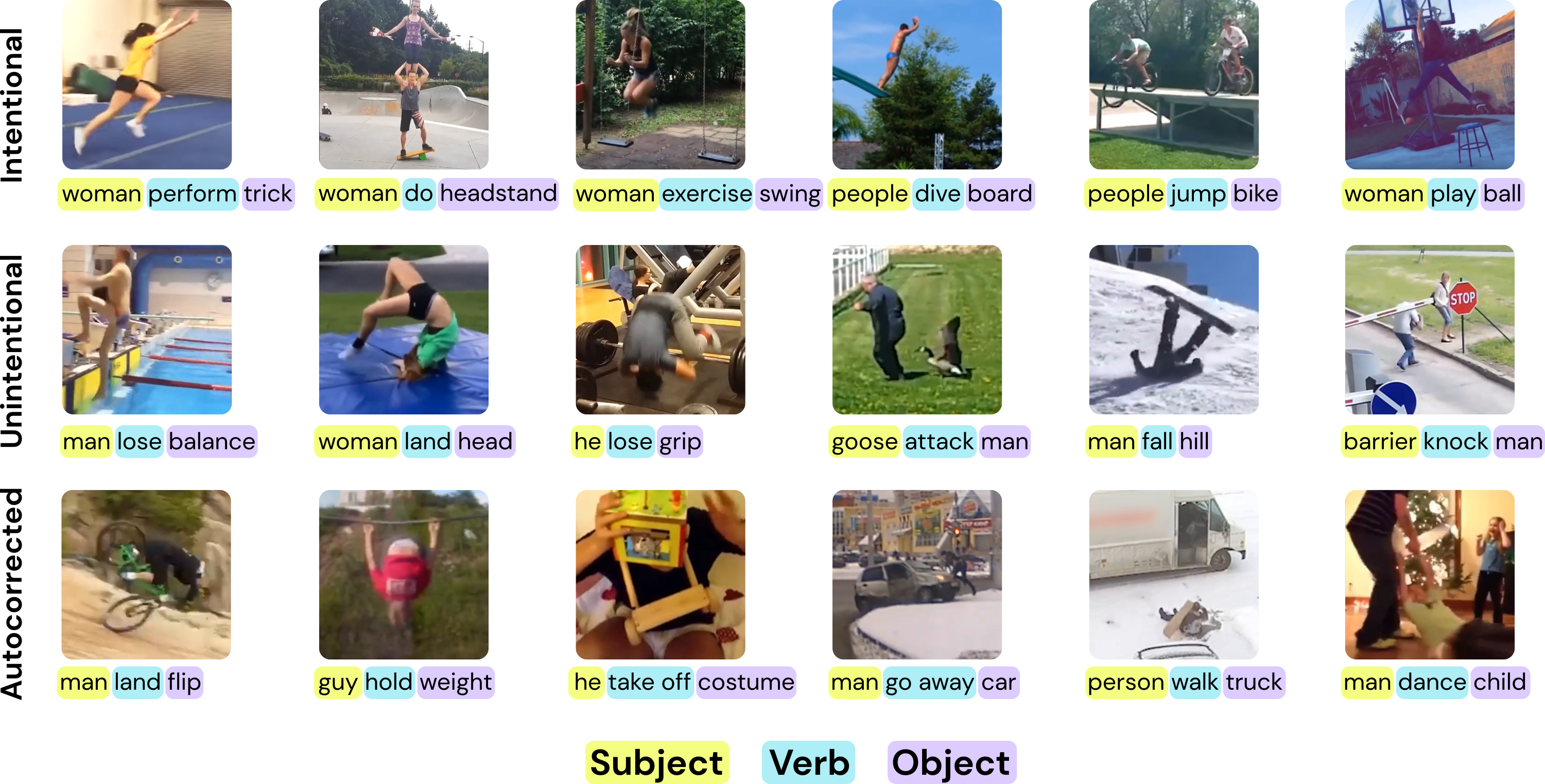}
% \vspace{-0.5em}
\caption{\textbf{Decoding the Trajectories:} We run our trained subject-verb-object decoder on different segments of Oops! videos. Row 1 shows clips of intentional action, and the trained decoder predicts the latent goal. Row 2 shows unintentional action, and the trained decoder now predicts failures instead. The final row also shows unintentional videos, but we run our auto-correction algorithm before predicting SVOs. The trained decoder returns to predicting goals, suggesting the auto-correct procedure shifts the failed trajectories towards successful ones.}
% After estimating the decoder, we read out triplets from different parts of videos. The first row shows intentional action, and the decoder predicts the goal. The second row shows unintentional action, and the decoder now predicts the failure instead. The final row shows unintentional videos that have been auto-corrected, and the decoder returns to predicting goals, suggesting the auto-correct procedure shifts the failed trajectories towards successful ones. \de{clarify experiment}
\label{fig:svos} 
%\vspace{-1em}
\end{figure*}
To solve this optimization problem, we use the iterative target class method \cite{kurakin2016adversarial}, which repeatedly runs the input through the model and modifies it in the direction of the desired loss. For every $\phi_i$ corresponding to a clip where action is unintentional, we repeat a gradient attack step towards the target $y^{\text{fail}}_i = 0$.
%until the condition $\hat{y}^{\text{intention}}_i = 0$ is met.
The complete update is:\footnote{We found $k_{max} = 25, \alpha = 0.03, \epsilon = 1$ to be reasonable values.}
\begin{align}
\phi^{k+1}_{t:T} = \ \texttt{clip} \left[\phi^k_{t:T} - \alpha \ \texttt{sign}\left(\nabla_{\phi_{t:T}} J\right), \phi_{t:T} \pm \epsilon\right]
\end{align}
where $\phi^0_{t:T} = \phi_{t:T}$. We repeat this process until the network is ``fooled'' into classifying the input as intentional action, for at most $k_\text{max}$ iterations or until $\arg\max \hat{y}^{\text{fail}}_i = 0$. Once the halting condition is satisfied, we run the modified $\phi^\prime$ vectors through the model, yielding a trajectory of corrected action $h^\prime$ that encodes successful completion of the goal. In other words, goals are the adversarial examples \cite{goodfellow2014explaining} of failed action -- instead of viewing adversarial examples as a bug, we view them as a feature \cite{NIPS2019_8307}.

 As a comparison, we implement a simple baseline where we linearly extrapolate the trajectory of observed intentional action: if the unintentional action in a sequence of clips $\{x_i\}_{i=0}^n$ begins at clip $j$, we extend the trajectory for a clip $x_k \in \{x_{j}, \ldots, x_n\}$ by setting $h_k = h_j + (k-j)\frac{h_j - h_0}{j}$. We find this baseline to outperform other naive ones such as the identity function (\ie leaving the representation untouched) and using the representation of the last moment before unintentional action.

Figure \ref{fig:autocorr_retrs} shows examples of nearest neighbor retrievals of the corrected latent vectors, computing over the Oops! and Kinetics test sets. Despite not training on Kinetics (i.e.\ on videos with completed goals), our representation can adjust video trajectories such that their nearest neighbors are goals being successfully executed. We also examine the effects of auto-correction on the frozen SVO decoder. Table \ref{tbl:ac_stats_eval_svo} shows these results. For decoders trained on all models, rankings of intentional action SVOs increase while those of unintentional SVOs decrease. However, the changes are greatest for our model. Figure \ref{fig:svos} visualizes the output of a frozen SVO decoder on auto-corrected actions, demonstrating the auto-correct process' ability to encode completed goals in its output trajectories.

\begin{figure*}[t]
\centering
\begin{subfigure}{.5\textwidth}
  \includegraphics[width=0.95\linewidth]{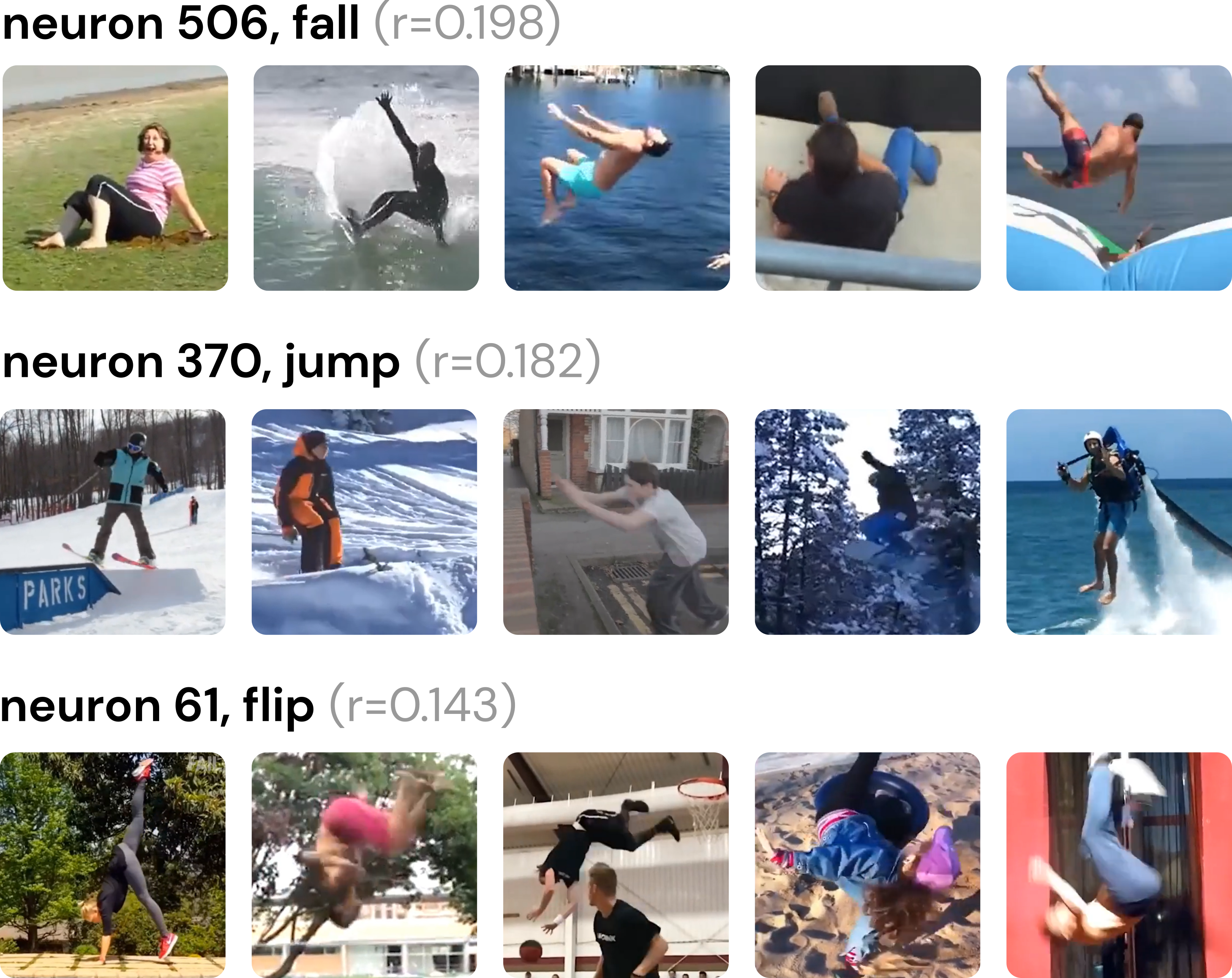}
  \caption{Top neuron-SVO correlations}
  \label{fig:neurons}
\end{subfigure}%
\begin{subfigure}{.5\textwidth}
  \centering
  \includegraphics[width=0.95\linewidth]{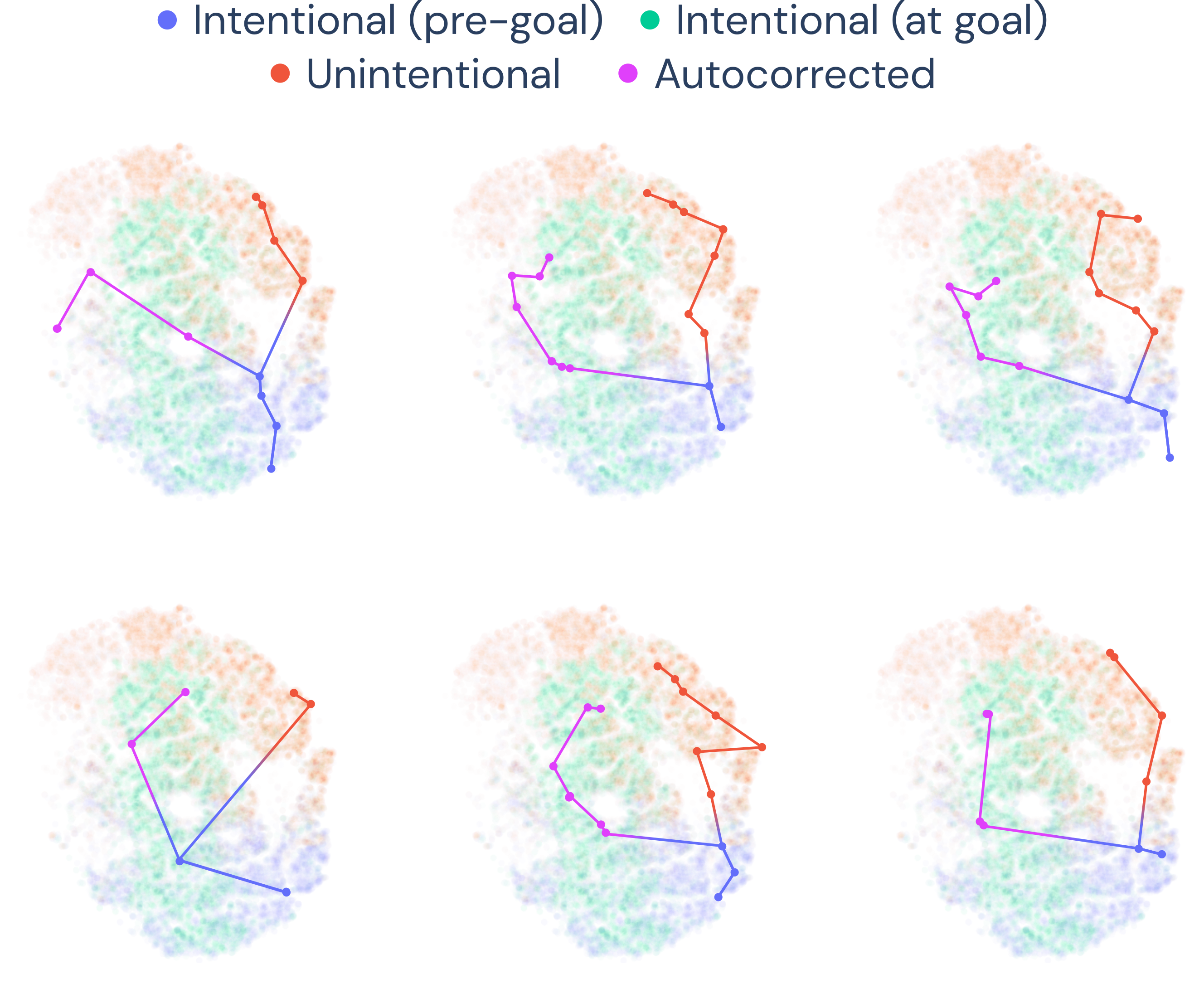}
  \caption{Trajectories in t-SNE}
  \label{fig:tsne}
\end{subfigure}
\caption{\textbf{Analyzing the Representation:} We probe the learned trajectories. (\textbf{a}) shows the neurons with highest correlation to the words in the SVO vocabulary, along with their top-5 retrieved clips. Neurons that detect intentions across a wide range of action and scene appear to emerge, despite only training with binary labels on the intentionality of action. (\textbf{b}) We show six randomly sampled video trajectories in t-SNE space, before and after auto-correct, superimposed over the embeddings for intentional and unintentional action. Visualizations suggest our approach tends to adjust unintentional action in the direction of successful, intentional action.}
\label{fig:test}
\end{figure*}

\begin{table}%[1][t]
\resizebox{\columnwidth}{!}{
\small
\begin{tabular}{l l | c c | c c}
\toprule
&& \multicolumn{2}{c|}{\textbf{Intentional SVO}} & \multicolumn{2}{c}{\textbf{Unintentional SVO}} \\
\textbf{Method} & \textbf{Features} & \textbf{$\mathbf{\Delta}$ R5} & \textbf{$\mathbf{\Delta}$ Rank} & \textbf{$\mathbf{\Delta}$ R5} & \textbf{$\mathbf{\Delta}$ Rank} \\ \midrule
 Adversarial & Ours & \textbf{+1.6} & \textbf{+15.8M} & \textbf{-3.3} & \textbf{-9.3M} \\ 
 & Kinetics \cite{carreira2017quo} & +0.4 & +0.3M & -0.3 & -1.2M \\
& 3D CNN \cite{oops} & +0.3 & +0.1M & -0.3 & -0.6M \\ \midrule
 Linearized & Ours & +0.6 & +1.0M & -0.5 & -1.7M \\ \bottomrule
\end{tabular}%
% \vspace{-2em} %
}
\caption{\textbf{Evaluating Autocorrection:}  We freeze the trained SVO decoder and run it on trajectories of unintentional action, before and after auto-correction. We run our algorithm based on adversarial attacks in various feature spaces as well as a linearization baseline. Using our algorithm, the frozen decoder more often predicts the ground truth goal SVO instead of the failure, indicating that our representation -- crucially trained on unintentional and intentional action -- captures the goals latent to video.}
\label{tbl:ac_stats_eval_svo}
%\end{wraptable}
% \vspace{-2em}
\end{table}
%\subsection{Correcting Unintentional Action}

\label{sec:ac_eval}
\subsection{Analysis of Learned Representation}

% We find that models require around 12 iterations on average to complete the auto-correct process. All models eventually manage to correct the around 93\% of the test set, suggesting that some videos are intrinsically not correctable by our process. 

% Interestingly, accuracy on predicting intentional action of {\em pre-failure} moments increases, as does temopral coherence prediction accuracy. We hypothesize this is due to the decreased entropy of semantically consistent intentional action.

%To establish a point of comparison, we implement a simple linearization baseline: if the unintentional action in a sequence of clips $\{x_i\}_{i=0}^n$ begins at clip $j$, we extend the trajectory for a clip $x_k \in \{x_{j}, \ldots, x_n\}$ by setting $h_k = h_j + (k-j)\frac{h_j - h_0}{j}$.  Our adversarial approach outperforms this baseline quantitatively in SVO evaluation (Table \ref{tbl:ac_stats_eval_svo}) and qualitatively in retrievals (Figure \ref{fig:autocorr_retrs}).

% \input{ac_tables}
% \de{TABLE: stats table - n iters, etc, with all models}
% \de{FIGURE: many retrieval examples}
% \de{discuss delta entailment loss and entail freq, relate to delta pre oops acc, success @ 50}
% \de{TABLE: AC SVO stats, with all models}

We finally probe the model's learned representation to analyze how trajectories are encoded.  We measure Spearman's rho correlation between the activation of neurons in the output vectors $h \in \mathbb{R}^{512}$ and words in the SVO vocabulary. Each video is an observation containing neuron activations and an indicator variable for whether each word is present in ground truth. Many neurons have significant correlation, and we show the top 3 in Figure \ref{fig:neurons}, along with the 5 clips that activate them most. These neurons appear to discover common intentions in the Oops! dataset, despite being trained without any labels other than the moment of onset of unintentional action. Note that the neurons are often invariant to action class and capture shared underlying intention. We also visualize trajectories of some videos using t-SNE (Figure \ref{fig:tsne}), before and after autocorrect. Our model often adjusts trajectories from unintentional action to the region of embedding space with Kinetics videos, shown in the figure as ``at goal" action.

We evaluate our model's ability to classify action intentionality, predict goals, and automatically correct unintentional action. We train from scratch using the Oops! dataset \cite{oops} as described above.

\section{Implementation Details}
To train our model, we randomly sample sequences of clips $\{x_i\}_{i=1}^{n}$, where each clip $x_i$ consists of $k=16$ frames at $r=16$ fps. In training, the length of these sequences $n$ is randomly drawn between $[n_{lo}, n_{hi}] = [6,10]$, so the model trains on video segments up to 10 seconds long (due to GPU memory limitations). Each clip is input to a 3D CNN $f_{cnn}$ (we use the R2+1D-18 architecture \cite{r2plus1d}) which gives a video token embedding $\phi_i = f_{cnn}(x_i) \in \mathbb{R}^{d}$, where $d=512$ is the hidden representation dimension. This is analogous to the word token embedding common in language modeling ({\em e.g.} \cite{BERT}), where we separately learn embeddings for special tokens used to delimit input sequences. 
% A small embedding matrix $W \in \mathbb{R}^{d \times 3}$ is maintained for three special tokens used to delimit input sequences: \texttt{[CLS]}, \texttt{[SEP]}, and \texttt{[PAD]}, with corresponding one hot vectors in $\mathbb{R}^3$ (borrowing from BERT).

In addition to video and special token embeddings, an additional function embeds each token's position in the input sequence, since the Transformer's attention-based computations do not otherwise encode input positions. This embedding is fixed to a combination of trigonometric functions as in \cite{transformer}, and is added to the CNN output. This allows the network to learn to generalize to unseen sequence lengths at test time, crucial to allow inference on very long videos (which would not fit in the GPU during training due to computational graph overhead). Input token embeddings are then fed to a 4-layer Transformer network with 8 attention heads per layer. For more details, please see Supplementary Material.

% The final embedding of a video clip $x^A_i$ belonging to video $A$ and at position $p$ is then $\phi_i = f_{cnn}(x_i) + f_{type}(vid_{A}) + f_{pos}(p)$. For special tokens, $\phi_i = W_i + f_{type}(special) + f_{pos}(p)$. Input to the whole model is either one or two video sequences (depending on whether input is split for the temporal coherence loss or not): {$ \ \texttt{[CLS]}_0 \ x_1, \ldots, x_{n} \ \texttt{[SEP]}_{n+1} \ $} or {$ \ \texttt{[CLS]}_0 \ x^A_1, \ldots, x^A_i \ \texttt{[SEP]}_{i+1} \ x^B_{i+2}, \ldots, x^B_{n+1} \ \texttt{[SEP]}_{n+2} \ $}, where $n \in [n_{lo}, n_{hi}]$, superscript denotes video ID ($A$ may or may not be equal to $B$), and subscript denotes overall position index. The transformer  then takes in embeddings $\phi$ and outputs hidden representations $h$ at each location. We run $h$ through a GELU \cite{hendrycks2016gaussian} before using in downstream tasks.

% We use the Adam optimizer with learning rate $3 \times 10^{-5}$, $\epsilon=10^{-4}$, and all other parameters standard. We accumulate gradients in batches of 8 before backward passes for training stability. We train using NVIDIA's Apex library for faster FP-16 training.

At test time, we feed entire videos through our model, sampled in continuous one-second intervals. If running auto-correct, we automatically split the model into two sequences at the clip where unintentional action is predicted to begin. Otherwise, we keep the entire video intact and represent it as a full trajectory.

\section{Conclusion}

We introduce an approach to learn about goals in video. By encoding action as a trajectory, we are able to perform several different tasks, such as decoding to categorical descriptions or manipulating the trajectory. Our experiments show that learning from failure examples, not just successful action, is crucial for learning rich visual representations of goals.  

\textbf{Acknowledgements:}
This research is based on work partially supported by the DARPA MCS program under Federal Agreement No.\ N660011924032, NSF NRI Award \#1925157, and an Amazon Research Gift. We thank NVidia for GPU donations. The views and conclusions contained herein are those of the authors and should not be interpreted as necessarily representing the official policies, either expressed or implied, of the U.S. Government.

{
\bibliographystyle{ieee_fullname}
\bibliography{main.bib}

\begin{thebibliography}{10}\itemsep=-1pt

\bibitem{athalye2017synthesizing}
Anish Athalye, Logan Engstrom, Andrew Ilyas, and Kevin Kwok.
\newblock Synthesizing robust adversarial examples.
\newblock {\em arXiv preprint arXiv:1707.07397}, 2017.

\bibitem{barresi1996intentional}
John Barresi and Chris Moore.
\newblock Intentional relations and social understanding.
\newblock {\em Behavioral and brain sciences}, 19(1):107--122, 1996.

\bibitem{kinetics}
Joao Carreira, Eric Noland, Andras Banki-Horvath, Chloe Hillier, and Andrew
  Zisserman.
\newblock A short note about kinetics-600.
\newblock {\em arXiv preprint arXiv:1808.01340}, 2018.

\bibitem{carreira2017quo}
Joao Carreira and Andrew Zisserman.
\newblock Quo vadis, action recognition? a new model and the kinetics dataset.
\newblock In {\em proceedings of the IEEE Conference on Computer Vision and
  Pattern Recognition}, pages 6299--6308, 2017.

\bibitem{colle2007childrens}
Livia Colle, Divide Mate, Marco Del~Giudice, Chris Ashwin, and Simon~Baron
  Cohen.
\newblock Childrens understanding of intentional vs. non-intentional action.
\newblock {\em Journal of Cognitive Science}, 8(1):39--68, 2007.

\bibitem{BERT}
Jacob Devlin, Ming-Wei Chang, Kenton Lee, and Kristina Toutanova.
\newblock Bert: Pre-training of deep bidirectional transformers for language
  understanding.
\newblock {\em arXiv preprint arXiv:1810.04805}, 2018.

\bibitem{doughty2018s}
Hazel Doughty, Dima Damen, and Walterio Mayol-Cuevas.
\newblock Who's better? who's best? pairwise deep ranking for skill
  determination.
\newblock In {\em Proceedings of the IEEE Conference on Computer Vision and
  Pattern Recognition}, pages 6057--6066, 2018.

\bibitem{oops}
Dave Epstein, Boyuan Chen, and Carl Vondrick.
\newblock Oops! predicting unintentional action in video.
\newblock {\em arXiv preprint arXiv:1911.11206}, 2019.

\bibitem{feichtenhofer2019slowfast}
Christoph Feichtenhofer, Haoqi Fan, Jitendra Malik, and Kaiming He.
\newblock Slowfast networks for video recognition.
\newblock In {\em Proceedings of the IEEE International Conference on Computer
  Vision}, pages 6202--6211, 2019.

\bibitem{fernando2017self}
Basura Fernando, Hakan Bilen, Efstratios Gavves, and Stephen Gould.
\newblock Self-supervised video representation learning with odd-one-out
  networks.
\newblock In {\em Proceedings of the IEEE conference on computer vision and
  pattern recognition}, pages 3636--3645, 2017.

\bibitem{gao2019listen}
Ruohan Gao, Tae-Hyun Oh, Kristen Grauman, and Lorenzo Torresani.
\newblock Listen to look: Action recognition by previewing audio.
\newblock {\em arXiv preprint arXiv:1912.04487}, 2019.

\bibitem{goodfellow2014explaining}
Ian~J Goodfellow, Jonathon Shlens, and Christian Szegedy.
\newblock Explaining and harnessing adversarial examples.
\newblock {\em arXiv preprint arXiv:1412.6572}, 2014.

\bibitem{gu2018ava}
Chunhui Gu, Chen Sun, David~A Ross, Carl Vondrick, Caroline Pantofaru, Yeqing
  Li, Sudheendra Vijayanarasimhan, George Toderici, Susanna Ricco, Rahul
  Sukthankar, et~al.
\newblock Ava: A video dataset of spatio-temporally localized atomic visual
  actions.
\newblock In {\em Proceedings of the IEEE Conference on Computer Vision and
  Pattern Recognition}, pages 6047--6056, 2018.

\bibitem{han2019video}
Tengda Han, Weidi Xie, and Andrew Zisserman.
\newblock Video representation learning by dense predictive coding.
\newblock In {\em Proceedings of the IEEE International Conference on Computer
  Vision Workshops}, pages 0--0, 2019.

\bibitem{han2020memory}
Tengda Han, Weidi Xie, and Andrew Zisserman.
\newblock Memory-augmented dense predictive coding for video representation
  learning.
\newblock {\em arXiv preprint arXiv:2008.01065}, 2020.

\bibitem{han2020self}
Tengda Han, Weidi Xie, and Andrew Zisserman.
\newblock Self-supervised co-training for video representation learning.
\newblock {\em Advances in Neural Information Processing Systems}, 33, 2020.

\bibitem{hara2018can}
Kensho Hara, Hirokatsu Kataoka, and Yutaka Satoh.
\newblock Can spatiotemporal 3d cnns retrace the history of 2d cnns and
  imagenet?
\newblock In {\em Proceedings of the IEEE conference on Computer Vision and
  Pattern Recognition}, pages 6546--6555, 2018.

\bibitem{hayes2017generating}
Jamie Hayes and George Danezis.
\newblock Generating steganographic images via adversarial training.
\newblock In {\em Advances in Neural Information Processing Systems}, pages
  1954--1963, 2017.

\bibitem{hays2007scene}
James Hays and Alexei~A Efros.
\newblock Scene completion using millions of photographs.
\newblock {\em ACM Transactions on Graphics (TOG)}, 26(3):4--es, 2007.

\bibitem{hochreiter1997long}
Sepp Hochreiter and J{\"u}rgen Schmidhuber.
\newblock Long short-term memory.
\newblock {\em Neural computation}, 9(8):1735--1780, 1997.

\bibitem{NIPS2019_8307}
Andrew Ilyas, Shibani Santurkar, Dimitris Tsipras, Logan Engstrom, Brandon
  Tran, and Aleksander Madry.
\newblock Adversarial examples are not bugs, they are features.
\newblock In H. Wallach, H. Larochelle, A. Beygelzimer, F. dAlch\'{e} Buc, E.
  Fox, and R. Garnett, editors, {\em Advances in Neural Information Processing
  Systems 32}, pages 125--136. Curran Associates, Inc., 2019.

\bibitem{jahanian2019steerability}
Ali Jahanian, Lucy Chai, and Phillip Isola.
\newblock On the''steerability" of generative adversarial networks.
\newblock {\em arXiv preprint arXiv:1907.07171}, 2019.

\bibitem{jayaraman2016slow}
Dinesh Jayaraman and Kristen Grauman.
\newblock Slow and steady feature analysis: higher order temporal coherence in
  video.
\newblock In {\em Proceedings of the IEEE Conference on Computer Vision and
  Pattern Recognition}, pages 3852--3861, 2016.

\bibitem{ji2019action}
Jingwei Ji, Ranjay Krishna, Li Fei-Fei, and Juan~Carlos Niebles.
\newblock Action genome: Actions as composition of spatio-temporal scene
  graphs.
\newblock {\em arXiv preprint arXiv:1912.06992}, 2019.

\bibitem{3dcnn}
Shuiwang Ji, Wei Xu, Ming Yang, and Kai Yu.
\newblock 3d convolutional neural networks for human action recognition.
\newblock {\em IEEE transactions on pattern analysis and machine intelligence},
  35(1):221--231, 2012.

\bibitem{jiang2019black}
Linxi Jiang, Xingjun Ma, Shaoxiang Chen, James Bailey, and Yu-Gang Jiang.
\newblock Black-box adversarial attacks on video recognition models.
\newblock In {\em Proceedings of the 27th ACM International Conference on
  Multimedia}, pages 864--872, 2019.

\bibitem{klaser2008spatio}
Alexander Klaser, Marcin Marsza{\l}ek, and Cordelia Schmid.
\newblock A spatio-temporal descriptor based on 3d-gradients.
\newblock 2008.

\bibitem{kurakin2016adversarial}
Alexey Kurakin, Ian Goodfellow, and Samy Bengio.
\newblock Adversarial examples in the physical world.
\newblock {\em arXiv preprint arXiv:1607.02533}, 2016.

\bibitem{laptev2005space}
Ivan Laptev.
\newblock On space-time interest points.
\newblock {\em International journal of computer vision}, 64(2-3):107--123,
  2005.

\bibitem{li2020ava}
Ang Li, Meghana Thotakuri, David~A Ross, Jo{\~a}o Carreira, Alexander
  Vostrikov, and Andrew Zisserman.
\newblock The ava-kinetics localized human actions video dataset.
\newblock {\em arXiv preprint arXiv:2005.00214}, 2020.

\bibitem{luvizon20182d}
Diogo~C Luvizon, David Picard, and Hedi Tabia.
\newblock 2d/3d pose estimation and action recognition using multitask deep
  learning.
\newblock In {\em Proceedings of the IEEE Conference on Computer Vision and
  Pattern Recognition}, pages 5137--5146, 2018.

\bibitem{meltzoff1995understanding}
Andrew~N Meltzoff.
\newblock Understanding the intentions of others: re-enactment of intended acts
  by 18-month-old children.
\newblock {\em Developmental psychology}, 31(5):838, 1995.

\bibitem{meltzoff1999toddlers}
Andrew~N Meltzoff, Alison Gopnik, and Betty~M Repacholi.
\newblock Toddlers' understanding of intentions, desires and emotions:
  Explorations of the dark ages.
\newblock 1999.

\bibitem{misra2016shuffle}
Ishan Misra, C~Lawrence Zitnick, and Martial Hebert.
\newblock Shuffle and learn: unsupervised learning using temporal order
  verification.
\newblock In {\em European Conference on Computer Vision}, pages 527--544.
  Springer, 2016.

\bibitem{miyato2016adversarial}
Takeru Miyato, Andrew~M Dai, and Ian Goodfellow.
\newblock Adversarial training methods for semi-supervised text classification.
\newblock {\em arXiv preprint arXiv:1605.07725}, 2016.

\bibitem{miyato2015distributional}
Takeru Miyato, Shin-ichi Maeda, Masanori Koyama, Ken Nakae, and Shin Ishii.
\newblock Distributional smoothing with virtual adversarial training.
\newblock {\em arXiv preprint arXiv:1507.00677}, 2015.

\bibitem{osadchy2017no}
Margarita Osadchy, Julio Hernandez-Castro, Stuart Gibson, Orr Dunkelman, and
  Daniel P{\'e}rez-Cabo.
\newblock No bot expects the deepcaptcha! introducing immutable adversarial
  examples, with applications to captcha generation.
\newblock {\em IEEE Transactions on Information Forensics and Security},
  12(11):2640--2653, 2017.

\bibitem{Parmar_2017_CVPR_Workshops}
Paritosh Parmar and Brendan Tran~Morris.
\newblock Learning to score olympic events.
\newblock In {\em The IEEE Conference on Computer Vision and Pattern
  Recognition (CVPR) Workshops}, July 2017.

\bibitem{pirsiavash2014parsing}
Hamed Pirsiavash and Deva Ramanan.
\newblock Parsing videos of actions with segmental grammars.
\newblock In {\em Proceedings of the IEEE conference on computer vision and
  pattern recognition}, pages 612--619, 2014.

\bibitem{pirsiavash2014assessing}
Hamed Pirsiavash, Carl Vondrick, and Antonio Torralba.
\newblock Assessing the quality of actions.
\newblock In {\em European Conference on Computer Vision}, pages 556--571.
  Springer, 2014.

\bibitem{she2019neuzz}
Dongdong She, Kexin Pei, Dave Epstein, Junfeng Yang, Baishakhi Ray, and Suman
  Jana.
\newblock Neuzz: Efficient fuzzing with neural program smoothing.
\newblock In {\em 2019 IEEE Symposium on Security and Privacy (SP)}, pages
  803--817. IEEE, 2019.

\bibitem{shultz1980development}
Thomas~R Shultz, Diane Wells, and Mario Sarda.
\newblock Development of the ability to distinguish intended actions from
  mistakes, reflexes, and passive movements.
\newblock {\em British Journal of Social and Clinical Psychology},
  19(4):301--310, 1980.

\bibitem{simonyan2013deep}
Karen Simonyan, Andrea Vedaldi, and Andrew Zisserman.
\newblock Deep inside convolutional networks: Visualising image classification
  models and saliency maps.
\newblock {\em arXiv preprint arXiv:1312.6034}, 2013.

\bibitem{simonyan2014two}
Karen Simonyan and Andrew Zisserman.
\newblock Two-stream convolutional networks for action recognition in videos.
\newblock In {\em Advances in neural information processing systems}, pages
  568--576, 2014.

\bibitem{sun2019contrastive}
Chen Sun, Fabien Baradel, Kevin Murphy, and Cordelia Schmid.
\newblock Contrastive bidirectional transformer for temporal representation
  learning.
\newblock {\em arXiv preprint arXiv:1906.05743}, 2019.

\bibitem{tomasello2009usage}
Michael Tomasello.
\newblock The usage-based theory of language acquisition.
\newblock In {\em The Cambridge handbook of child language}, pages 69--87.
  Cambridge Univ. Press, 2009.

\bibitem{tomasello2005understanding}
Michael Tomasello, Malinda Carpenter, Josep Call, Tanya Behne, and Henrike
  Moll.
\newblock Understanding and sharing intentions: The origins of cultural
  cognition.
\newblock {\em Behavioral and brain sciences}, 28(5):675--691, 2005.

\bibitem{r2plus1d}
Du Tran, Heng Wang, Lorenzo Torresani, Jamie Ray, Yann LeCun, and Manohar
  Paluri.
\newblock A closer look at spatiotemporal convolutions for action recognition.
\newblock In {\em Proceedings of the IEEE conference on Computer Vision and
  Pattern Recognition}, pages 6450--6459, 2018.

\bibitem{transformer}
Ashish Vaswani, Noam Shazeer, Niki Parmar, Jakob Uszkoreit, Llion Jones,
  Aidan~N Gomez, {\L}ukasz Kaiser, and Illia Polosukhin.
\newblock Attention is all you need.
\newblock In {\em Advances in neural information processing systems}, pages
  5998--6008, 2017.

\bibitem{von2003captcha}
Luis Von~Ahn, Manuel Blum, Nicholas~J Hopper, and John Langford.
\newblock Captcha: Using hard ai problems for security.
\newblock In {\em International Conference on the Theory and Applications of
  Cryptographic Techniques}, pages 294--311. Springer, 2003.

\bibitem{wang2011action}
Heng Wang, Alexander Kl{\"a}ser, Cordelia Schmid, and Cheng-Lin Liu.
\newblock Action recognition by dense trajectories.
\newblock In {\em CVPR 2011}, pages 3169--3176. IEEE, 2011.

\bibitem{wei2018learning}
Donglai Wei, Joseph~J Lim, Andrew Zisserman, and William~T Freeman.
\newblock Learning and using the arrow of time.
\newblock In {\em Proceedings of the IEEE Conference on Computer Vision and
  Pattern Recognition}, pages 8052--8060, 2018.

\bibitem{wei2018transferable}
Xingxing Wei, Siyuan Liang, Ning Chen, and Xiaochun Cao.
\newblock Transferable adversarial attacks for image and video object
  detection.
\newblock {\em arXiv preprint arXiv:1811.12641}, 2018.

\bibitem{woodward2009infants}
Amanda~L Woodward.
\newblock Infants' grasp of others' intentions.
\newblock {\em Current directions in psychological science}, 18(1):53--57,
  2009.

\bibitem{woodward2009emergence}
Amanda~L Woodward, Jessica~A Sommerville, Sarah Gerson, Annette~ME Henderson,
  and Jennifer Buresh.
\newblock The emergence of intention attribution in infancy.
\newblock {\em Psychology of learning and motivation}, 51:187--222, 2009.

\bibitem{woodward2001infants}
Amanda~L Woodward, Jessica~A Sommerville, and Jose~J Guajardo.
\newblock How infants make sense of intentional action.
\newblock {\em Intentions and intentionality: Foundations of social cognition},
  pages 149--169, 2001.

\bibitem{wu2017sampling}
Chao-Yuan Wu, R Manmatha, Alexander~J Smola, and Philipp Krahenbuhl.
\newblock Sampling matters in deep embedding learning.
\newblock In {\em Proceedings of the IEEE International Conference on Computer
  Vision}, pages 2840--2848, 2017.

\bibitem{wu2020comprehensive}
Zonghan Wu, Shirui Pan, Fengwen Chen, Guodong Long, Chengqi Zhang, and S~Yu
  Philip.
\newblock A comprehensive survey on graph neural networks.
\newblock {\em IEEE Transactions on Neural Networks and Learning Systems},
  2020.

\bibitem{xu2017r}
Huijuan Xu, Abir Das, and Kate Saenko.
\newblock R-c3d: Region convolutional 3d network for temporal activity
  detection.
\newblock In {\em Proceedings of the IEEE international conference on computer
  vision}, pages 5783--5792, 2017.

\bibitem{zhou2018invisible}
Zhe Zhou, Di Tang, Xiaofeng Wang, Weili Han, Xiangyu Liu, and Kehuan Zhang.
\newblock Invisible mask: Practical attacks on face recognition with infrared.
\newblock {\em arXiv preprint arXiv:1803.04683}, 2018.

\bibitem{zhu2018hidden}
Jiren Zhu, Russell Kaplan, Justin Johnson, and Li Fei-Fei.
\newblock Hidden: Hiding data with deep networks.
\newblock In {\em Proceedings of the European Conference on Computer Vision
  (ECCV)}, pages 657--672, 2018.

\end{thebibliography}
}
\typeout{get arXiv to do 4 passes: Label(s) may have changed. Rerun}
\end{document}